\documentclass[10pt, a4paper]{article}
\usepackage{lrec}
\usepackage{graphicx}
\usepackage{tabularx}
\usepackage{soul}
\usepackage{multirow}

\usepackage{epstopdf}
\usepackage[latin1,utf8]{inputenc}

\usepackage{hyperref}
\usepackage{xstring}

\title{A Scientific Information Extraction Dataset for Nature Inspired Engineering}

\name{Ruben Kruiper$^{*}$, Julian F.V. Vincent, Jessica Chen-Burger, \\ 
{\bf \large Marc P.Y. Desmulliez, Ioannis Konstas}}

\address{Heriot-Watt University, Riccarton Campus \\
        Edinburgh, United Kingdom EH14 4AS\\
        $^{*}$Corresponding author: rk22@hw.ac.uk
        }

\abstract{
Nature has inspired various ground-breaking technological developments in applications ranging from robotics to aerospace engineering and the manufacturing of medical devices. However, accessing the information captured in scientific biology texts is a time-consuming and hard task that requires domain-specific knowledge. Improving access for outsiders can help interdisciplinary research like Nature Inspired Engineering. This paper describes a dataset of 1,500 manually-annotated sentences that express domain-independent relations between central concepts in a scientific biology text, such as trade-offs and correlations. The arguments of these relations can be Multi Word Expressions and have been annotated with modifying phrases to form non-projective graphs. The dataset allows for training and evaluating Relation Extraction algorithms that aim for coarse-grained typing of scientific biological documents, enabling a high-level filter for engineers. \\ 
\newline 
}

\begin{document}
\maketitleabstract
\section{Introduction}

Discovering relevant scientific literature is a hard and time-consuming task that often requires domain expertise \cite{Alper2004,El-Arini2011,Pain2017}. This difficulty can lead to barriers in inter-disciplinary fields of study, where an expert in a target domain is often a novice in the source domain \cite{Carr2018}. A specific example of an inter-disciplinary field of study is Nature Inspired Engineering, also known as biomimetics \cite{Kruiper2016}. A well known example of biomimetics is Velcro\textsuperscript{\textregistered} \cite{Velcro1955}, an `\textit{adhesion}' method that is inspired by nature \-- specifically the burrs of burdock plants. In biomimetics an engineer is interested in learning from nature to solve a problem at hand \cite{Hoeller2016}. It has been shown that identifying, selecting and understanding relevant biological information is challenging for engineers \cite{Vattam2013a}. Information Extraction (IE) systems can partially ease these challenges by capturing the central concepts and relations in a text \cite{Augenstein2017,Gabor2018}.

\begin{figure}[t]
    \centering 
    \includegraphics[width=1\linewidth]{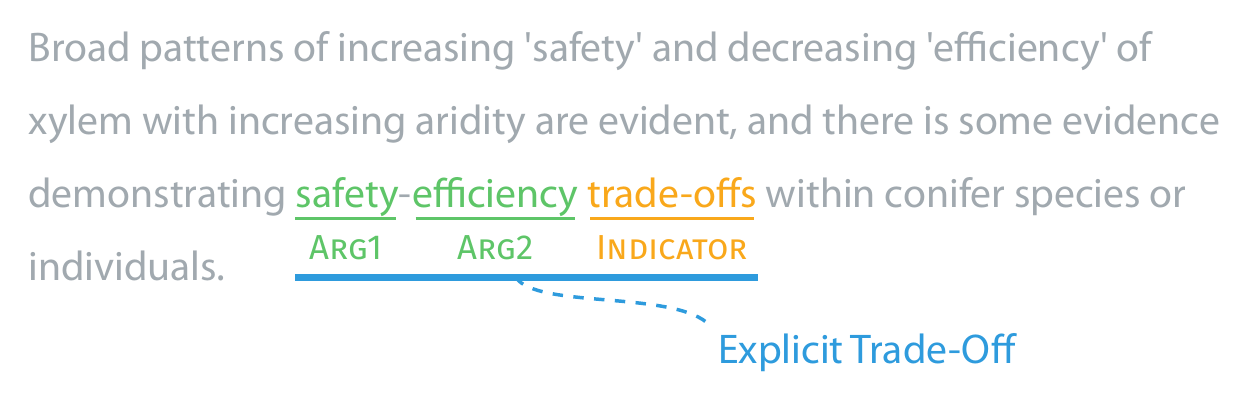}
    \caption{Example of a \textsc{Trade-Off} relation between domain-independent abstract concepts, denoted as \protect\textsc{Arg1} and \protect\textsc{Arg2}, extracted from \protect\cite{Burgess2006}.}
    \label{fig:tradeoff_example}
\end{figure}

A major issue in inter-disciplinary research is that search terms from a source domain may not retrieve relevant results in a target domain. As an example, in the biological domain the term `\textit{bleaching}' can refer to a separation process between the retina and opsin in vertebrate eyes \cite{Grimm2000}. A non-biologist may expect `\textit{bleaching}' to refer to \textit{cleaning, sterilizing}, or \textit{whitening} \cite{Nagel2014a}. As a result, terminology from the engineering domain does not always provide a good starting point for the identification of relevant biological information \cite{Fayemi2015,Kruiper2017}. A more appropriate approach that allows for cross-domain search, without relying so much on domain-specific semantics, focuses on \textsc{Trade-Off} relations \cite{Vincent2016,Kruiper2018}. In biology, trade-offs express how fitness is constrained by functional requirements that are mutually exclusive, e.g., because they share a limited resource \cite{Agrawal2010}. Figure~\ref{fig:tradeoff_example} provides an example of a trade-off between `\textit{safety}' and `\textit{efficiency}'. Trade-offs are important drivers for adaptation and speciation, and underpin much of the research in various biology sub-domains \cite{Stearns1989,Garland2014,Ferenci2016}. 
\textsc{Trade-Off} relations are of interest to biomimetics because they are able to capture the problem space of a full-text biology document in highly abstract terms. Analyzing these trade-offs enables a domain-independent document classification. Crucially, trade-offs can direct an engineer to the mechanisms that biological systems employ to manipulate such problem spaces. 

In this work we present the Focused Open Biology Information Extraction (\textsc{FOBIE}) dataset. \textsc{FOBIE} comprises 1500 manually-annotated sentences taken from full-text scientific biology documents. The dataset enables training and evaluation of Relation Extraction (RE) tools that extract trade-offs from scientific biology texts. Each sentence is annotated with a non-projective graph of \textit{n}-ary relations between trigger words, argument phrases and modifying phrases. The relations are domain-independent and the argument phrases contain concepts that are central to the text. Most of these phrases are not available in standard knowledge bases, so they cannot be learned through distant supervision. This paper provides a comparison of \textsc{FOBIE} with regards to existing datasets for scientific Information Extraction (IE), a description of the collection and annotation processes, and the results of a strong scientific IE baseline. 

\section{Related work}
\subsection{Computer-Aided biomimetics}
\label{ssec:CAB}
Biomimetics is an engineering problem solving process during which one draws on analogous biological solutions. While biomimetics has led to the development of leap-frog innovations, the process remains adventitious. A major issue is that engineers know little biology or few details of plants or animals. Therefore, engineers have trouble identifying, selecting and understanding relevant biological information \cite{Vattam2011a,Vattam2013b}. To overcome these challenges various computational tools have been developed, so-called Computer-Aided biomimetics (CAB) tools. Their biggest limitation is that they rely heavily on cross-domain mapping of information such as functional similarities, e.g., \cite{Vandevenne2012,Vandevenne2016b,Shu2014,Cheong2014,Rugaber2016,Zhao2019}. However, automatically identifying cross-domain relational similarities, requires substantial reasoning beyond what is currently computationally feasible.


Two caveats are lexical variation and polysemy of concepts across domains \cite{Augenstein2017}. Thirdly, the biological and engineering domains are fundamentally different. Biological systems develop through evolution and natural selection, while engineering is based on conscious decision-making to meet functional requirements \cite{Vincent2006}. Consider again the example of `\textit{bleaching}': the purely biological term does not carry a notion of teleology, in contrast to the non-biologist interpretations. The active verbs that are associated to \textit{engineering functions} are not always used in biological texts \cite{Kaiser2012,Kaiser2014}. As a result, each function of interest requires a separate classifier \cite{Glier2013} or its own set of extraction rules that work for a specific domain of texts \cite{Etzioni2007,Christensen2011}. Training data for each function would accordingly require annotation by a domain expert \cite{Luan2018a}.

Instead, this work aims to extract trade-off relations that are central to many biological texts and capture information at an abstraction level that is domain-independent \cite{Vincent2016}. These trade-offs, and the abstract concepts captured in their arguments, provide an initial filter for information retrieval and support for within-domain search for biological information \cite{Kruiper2018}.


\subsection{Scientific Information Extraction}
\label{ssec:relation_extraction}
Information Extraction (IE) from scientific text can (1) improve access to scientific information, beyond the possibilities of standard search engines \cite{Gabor2018,Gupta2011}, (2) provide valuable insight into research areas \cite{Tsai2013,Luan2018a}, (3) enable to quickly learn facts on unknown concepts, as extractions can provide a summary view for readers \cite{Mausam2016}, and (4) augment existing Knowledge Bases (KB) from unlabeled text \cite{Quirk2017}. However, annotating scientific text is non-trivial, as it requires domain-specific knowledge from experts. Large-scale crowdsourcing \cite{Tratz2019} or human-in-the-loop \cite{He2016} efforts can be unreliable for such tasks. 

To alleviate the manual labeling of \textsc{FOBIE} by a domain expert, we developed a simple Rule-Based System (RBS). Like traditional Relation Extraction (RE) systems this RBS relies on matching specific words to find a \textsc{Trade-Off} or similar relation \cite{Sarawagi2007}. Similar to Open IE (OIE) systems it relies on unlexicalized grammatical structures, e.g., syntactic patterns, to determine the argument phrases \cite{Etzioni2011}. It is expected that this enables flexibility of extracting trade-offs from a variety of domains \cite{Etzioni2005,Banko2007}. However, the handcrafted or learned syntactic patterns rely on how well input sentences are handled \cite{lechelle2018,Niklaus2018}. Sources of issues may include the length and complexity of sentences, the use of pronouns as subjects, the use of abbreviations as adjectives, handling prepositional phrases, and dealing with co-reference \cite{Schneider2017,Groth2018}. Machine learning approaches can improve recall for the long tail of patterns that will have to be identified \cite{Mausam2016}.



\medskip
By manually relabeling the output of our RBS system we rectify errors in argument boundaries and determine the correct relation label. Only few similar datasets exist for scientific IE:

\paragraph{\textsc{ScienceIE}} Semeval 2017 task 10 introduced the \textsc{ScienceIE} dataset, consisting of 500 paragraphs taken from full-text scientific documents in the domains of Computer Science, Material Science and Physics journal articles \cite{Augenstein2017}. It contains annotations of keyphrases that are classified as \textit{materials}, \textit{processes} or \textit{tasks}. Furthermore, hyponym- and synonym-relations between the keyphrases are captured.

\paragraph{\textsc{Semeval 2018}} The manually annotated Semeval 2018 task 7 dataset contains 6 relations types that are noted to occur regularly in scientific abstracts; \textit{usage, result, model, part-whole, topic} and \textit{comparison} \cite{Gabor2018}. It contains 500 abstracts from the domain of Computational Linguistics and draws on the \textsc{ACL RD-TEC 2.0} corpus \cite{Qasemizadeh2016} for entity annotation; \textit{technology and method, tool and library, language resource, language resource product, measures and measurements, models} and \textit{other}. Augenstein \textit{et al.} \shortcite{Augenstein2017a} found that the \textsc{ScienceIE} dataset contains a significantly higher proportion of long keyphrases in comparison to the \textsc{ACL RD-TEC 2.0} corpus. This is likely due to the different characteristics of sentences taken from abstracts and those in the main body. 

\paragraph{\textsc{SciERC}} The \textsc{SciERC} dataset consists of 500 abstracts taken from Artificial Intelligence conference and workshop proceedings \cite{Luan2018a}. It extends the entity and relation types of the Semeval 2018 task 7 dataset; the relation types are \textit{used-for, feature-of, hyponym-of, part-of, compare, conjunction} and \textit{corefence}, and the entity types are \textit{task, method, evaluation metric, material, other scientific terms}, and \textit{generic terms} that include anaphora and cataphora that are annotated for corefence resolution. 

\begin{table}[ht]
    \small
    \centering
    \begin{tabular}{c|cccc}
         \#     & FOBIE & \textsc{SciERC} & \textsc{ScienceIE} & SE '18  \\ \hline
         Arguments      & 5835  & 8089  & 9946  & 7483  \\
         Relations      & 4788  & 4716  & 672   & 1595  \\
         Rel/doc        & 3.1*  & 9.4   & 1.3   & 3.2  \\
    \end{tabular}
    \caption{Amount of arguments, relations and relations per instance for FOBIE, \textsc{SciERC}, \textsc{ScienceIE} and SemEval 2018 task 7. Rel/doc stands for relations per sentence* for FOBIE (per abstract or paragraph for the other datasets).}
    \label{tab:datasets_comparison}
\end{table}{}

\begin{figure*}
    \centering
    \includegraphics[width=1\linewidth]{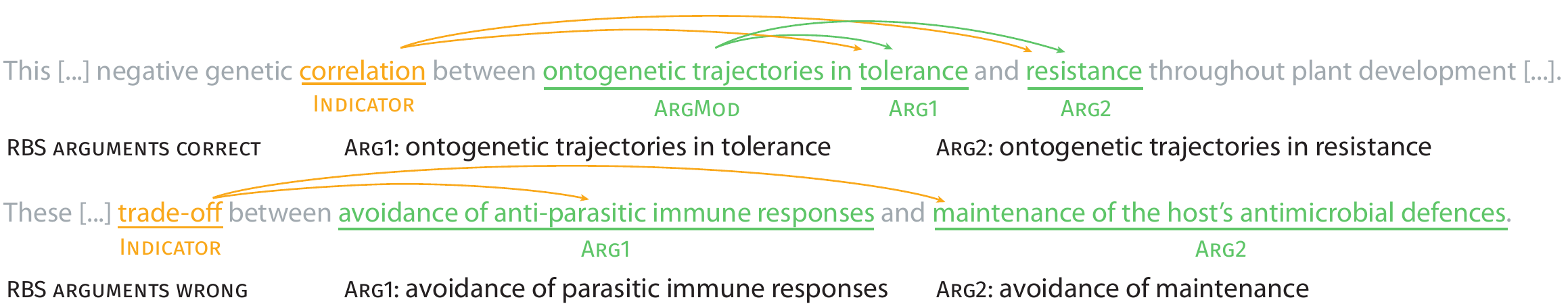}
    \caption{Two examples of sentences that express a trade-off. The indicator denotes the trigger word that connects two trade-off arguments \-- \textsc{Arg1} and \textsc{Arg2}. \textsc{ArgMod} denotes a modifying phrase.}
    \label{fig:RBS_output}
\end{figure*}


\medskip
In this work we introduce the \textsc{FOBIE} dataset that targets sentences taken from full-text scientific publications in the Biology domain. Table~\ref{tab:datasets_comparison} provides an overview of the sizes of \textsc{FOBIE}, \textsc{SciERC}, \textsc{ScienceIE} and the Semeval 2018 dataset. Similar to \textsc{ScienceIE} the focus lies on extracting keyphrases, rather than entities. Different from the previously described datasets, the relation arguments are not classified into a type, also see section \ref{sec:datasets} \textsc{FOBIE} enables the training of IE systems that extract both a 
narrow set of relations between argument phrases and an unbounded set of modifying relations.


\begin{figure*}
    \centering
    \includegraphics[width=1\linewidth]{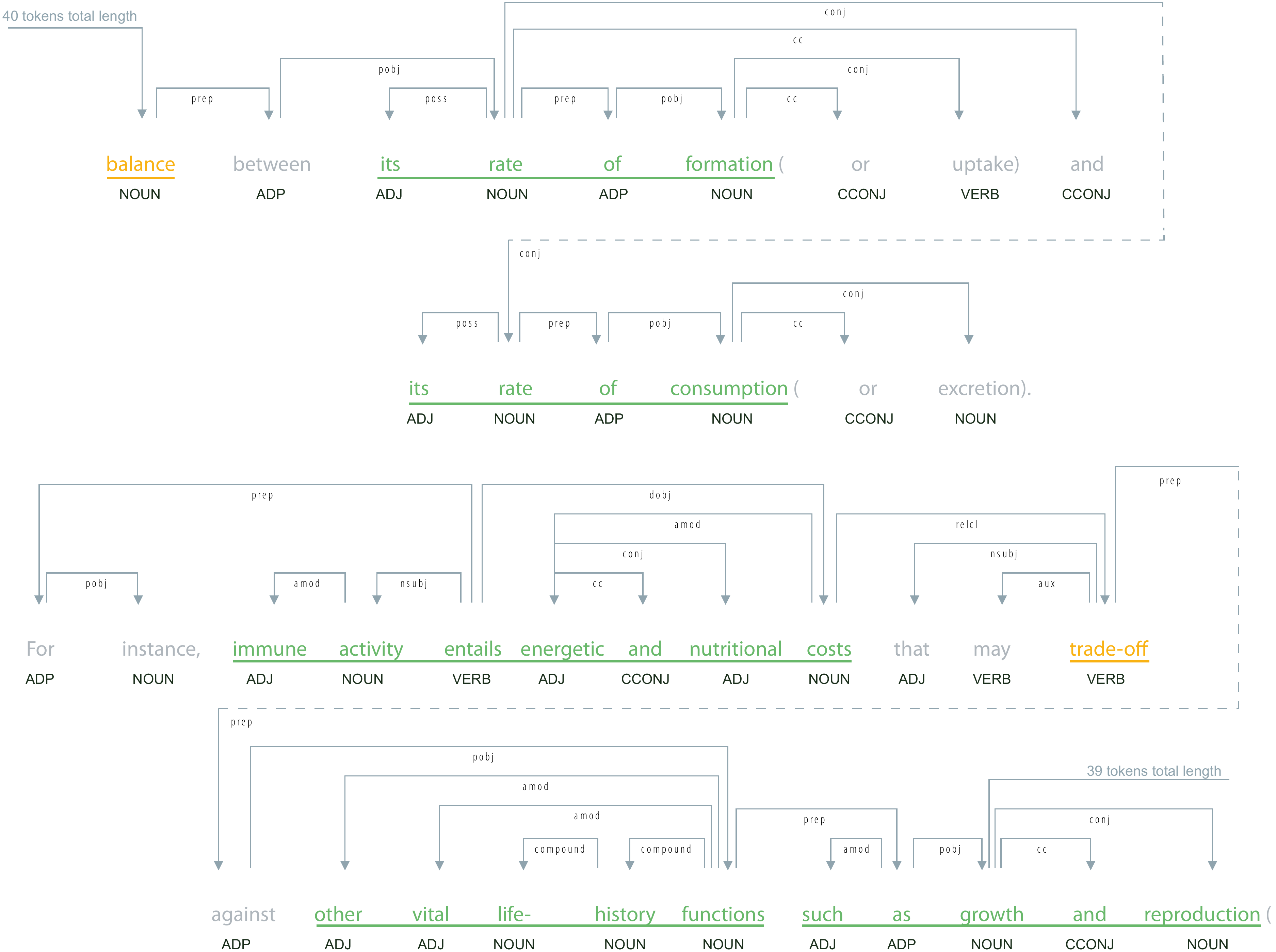}
    \caption{Dependency parses of two sentences that are near the average length of 37.81 tokens. The \textsc{Trade-Off} indicators and arguments have been underlined as in Figure~\ref{fig:RBS_output}, as well as the single argument modifier \textit{"such as [...]"}. Defining extraction rules that can traverse a large variety of dependency trees correctly is a time-consuming and hard task that requires expertise in both the linguistics and biology domains.
    }
    \label{fig:Dependency_examples}
\end{figure*}

\section{The FOBIE dataset} 
We collected a corpus of 10,000 open-access papers from the Journal of Experimental Biology (JEB) and three BioMed Central (BMC) journals: Biology, Systems Biology and Evolutionary Biology. We only retained the abstract, introduction, results, discussion and conclusion section. We used spaCy\footnote{https://spacy.io/} to split the texts into sentences and tokens, and computed dependency parses\footnote{spaCy parse accuracy on the Penn Treebank corpus is 94.48.}. 

\subsection{Rule-Based System}
A set of 200 papers was analyzed to determine the types of verbatim expressions that indicate a \textsc{Trade-Off} relation \cite{Kruiper2018}. The paper subjects ranged from cell biology to bio-mechanics and trigger words include \textit{association, balance, conflict, correlation, compromise, interaction, interplay, optimization, ratio, relationship} and \textit{trade-off}. We stem these trigger words and scan the sentences of the documents in our corpus for their presence. If a trigger word is found, we traverse the dependency parse tree to identify sub-trees that comprise argument phrases. 

The output of the RBS system indicates that it has difficulty dealing with negation, and that complex long sentences indeed lead to errors in extractions. Figure~\ref{fig:RBS_output} shows two sentences that express a \textit{trade-off}, where each sentence deals differently with phrase attachment and argument boundaries. Variance in argument and modifier arrangement complicate the definition of extraction rules that rely on a dependency parse tree. Therefore, an accurate RBS for the extraction of trade-offs from scientific text requires detailed extraction rules and exceptions. Figure~\ref{fig:Dependency_examples} shows two examples of partial dependency parses taken from sentences that are near the average length of $\pm$38 tokens. The main source of errors is due to the extraction rules themselves struggling with the large variety of verbatim trade-off expressions (e.g., `\textit{negative correlation}', `\textit{balance}'), rather than the quality of the dependency parser output. Improving the extraction rules of the simple RBS is expected to be a time-consuming and complex task. Very often the presence of a trigger word does not automatically lead to the presence of a trade-off. Determining whether the sentence expresses a trade-off requires additional reasoning and often domain-specific knowledge. Considering the large variety of syntactic structures that \textit{do} indicate a trade-off, as well as the required expertise in biology, a RBS is not suited for our task. Instead, the output of our simple RBS is used to speed up manual annotation.

\subsection{Dataset annotation} \label{sec:datasets}
Out of the 10,000 documents, we retain only the sentences that were annotated with a \textsc{Trade-Off} relation by the RBS. Using the BRAT\footnote{https://brat.nlplab.org/} interface a biology expert manually corrected argument boundaries and relation types \cite{Stenetorp2012}. During manual annotation we correct the relation label for trigger words, handle negation and identify the boundaries of argument and modifier phrases. We annotate binary relations that constitute non-projective graphs of one or more \textit{n}-ary relations in a sentence. Each binary relation is a triple ${<}governor, relation, dependent{>}$ where:
\begin{itemize}
    \item $governor$ is either a trigger word or a modifying phrase.
    \item $relation$ indicates the type of relation -- \textsc{Trade-off}, \textsc{Argument-Modifier} or \textsc{Not-a-Trade-off}.
    \item $dependent$ is an argument phrase.
\end{itemize}

Three relation types were used: \textsc{Trade-off}, \textsc{Argument-Modifier} and \textsc{Not-a-Trade-off}. The latter relation is used to indicate that a trigger word does not express a trade-off. We retain these annotations because their expressions are syntactically similar and, therefore, can provide useful training signal as negative samples. Furthermore, the trigger words to express these \textsc{Not-A-TradeOff} relations may be contiguous with the trigger words that do express a \textsc{Trade-off}.
An example is a sentence that expresses a correlation, which often denotes a positive correlation between two arguments while a \textsc{Trade-off} entails a negative correlation.

Many argument phrases are found to be nested and can be broken down into an argument and a modifier, e.g., `\textit{ontogenetic trajectories}' modifies `\textit{tolerance}' and `\textit{resistance}' in Figure~\ref{fig:RBS_output}. We do not indicate specific types of modification, such as temporal expressions or phrases that indicate a location, used in Semantic Role Labeling (SRL). The reason is that trigger words can be either nouns or verbs, e.g., `\textit{balance}/NOUN' and `\textit{trade-off}/VERB' in Figure~\ref{fig:Dependency_examples}, hence the classes of modifier labels found in common SRL frameworks such as Propbank \cite{Palmer2005} are not always appropriate. Instead, as a general rule we include the words that indicate the type of modifier during annotation of modifying phrases. 

We adopt the heuristic that prepositional phrases (PPs) heading a coordinating clause, are treated as modifying phrases when they apply to word-level arguments of a trade-off, as in the top example in Figure~\ref{fig:RBS_output}. In the case of coordination of PPs that contain arguments of a trade-off, we consider each PP as a whole argument, as in the bottom example in Figure~\ref{fig:RBS_output}. Similarly, when nested phrases in an argument can be distinctly separated by punctuation, such phrases are treated as an argument-modifier pair. We do not annotate phrases that modify the relation directly, e.g., `\textit{throughout plant development}' modifies `\textit{correlation}' in the top of Figure~\ref{fig:RBS_output}. Regarding the direction of relations, the trigger words and modifying phrases are treated as governors of a relation.

Nevertheless, determining exact rules for the boundaries of arguments was found to be challenging. As an example, consider the second \textsc{Trade-Off} relation in Figure~\ref{fig:Dependency_examples}. In this sentence, `\textit{immune activity}' is a `\textit{life-history function}' that trades off with `\textit{growth}' and `\textit{reproduction}', two other examples of `\textit{vital life-history functions}'. Specifically, `\textit{energetic and nutritional costs}' need to be shared between these `\textit{life-history functions}'. The first argument `\textit{immune activity [...] costs}' is treated as a single phrase following the heuristics above. This verb phrase could also be split up, leaving the informative phrase `\textit{entails energetic and nutritional costs}' outside of the annotation scope. In such ambiguous cases, we opted for always annotating the longest possible phrase spans, in order to capture more comprehensively the information in the sentence.

\begin{table}[tp]  
\small
\begin{tabular}{l|lll|l} 
                                  & Train & Dev     & Test  & Total     \\ \hline
\# Sentences                      & 1248  & 150     & 150   & 1548    \\
Avg. sent. length                 & 37.42 & 38.91   & 40.02 & 37.81   \\
\% of sents $\geq$ 25 tokens      & 82.21 &  85.33  & 83.33 & 82.62 \\\hline
\textbf{Relations}: & & & &\\
- \textsc{Trade-Off}              & 639   & 54      & 72    & 765     \\
- \textsc{Not-a-Trade-Off}        & 2004  & 258     & 240   & 2502    \\
- \textsc{Arg-Modifier}           & 1247  & 142     & 132   & 1521    \\\hline
Triggers                       & 1292  & 155        & 153   & 1600    \\
Keyphrases                     & 3436  & 401        & 398    & 4235    \\
Keyphrases w/ multiple rel's  & 1600 & 188          & 163   & 1951 \\
Spans                          & 4728  & 556        & 551   & 5835    \\
Max triggers/sent              & 2      & 2         & 2         & \\ 
Max spans/sent                 & 9     & 8         & 8         &  \\
Unique spans                   & 3643 \\
Unique triggers                & 41 \\

\# single-word keyphrases       & \multicolumn{3}{l}{864 (20.4\%)}  \\
Avg. tokens per keyphrase       & 3.46 \\
\end{tabular}
\caption{The aggregated statistics for \textsc{FOBIE}.}
\label{tab:dataset_description}
\end{table}
\vspace{2mm}

\begin{table}[tp]
\centering \small
\begin{tabular}{llc@{$\;$}|c@{$\;$}|cc@{$\;$}|c@{$\;$}|c}
    && \multicolumn{3}{c}{Relations} & \multicolumn{3}{c}{Boundaries} \\ 
    &&  \multicolumn{1}{c}{P}&\multicolumn{1}{c}{R}&\multicolumn{1}{c}{F1}& \multicolumn{1}{c}{P}&\multicolumn{1}{c}{R}&\multicolumn{1}{c}{F1}\\ \hline 
\multirow{2}{*}{\textbf{\textsc{RBS}}}& \textit{dev}  & \multicolumn{1}{c}{--}&\multicolumn{1}{c}{--}&\multicolumn{1}{c}{--}  & 45.17 & 35.84 & 39.97 \\
    & \textit{test}   & \multicolumn{1}{c}{--}&\multicolumn{1}{c}{--}&\multicolumn{1}{c}{--}   & 44.31 & 35.32 & 39.31 \\ \hline 
\multirow{2}{*}{\textbf{\textsc{SciIE}}}& \textit{dev} & 69.60 & 69.60 & 69.60  & 84.50 & 80.40 & 82.40\\
    & \textit{test}   & 68.28 & 69.82 & 69.04  & 84.26 & 79.67 & 81.90    
\end{tabular}
\caption{Overview of results on the development (dev) and test set of \textsc{FOBIE} for the \textsc{RBS} and \textsc{SciIE}. The relations columns refers to the Relation Extraction (RE) setting. The \textsc{RBS} does not classify relations and is, therefore, not evaluated on this task. Boundaries refers to determining the correct boundaries of arguments and trigger words.}
\label{tab:results}
\end{table}

\begin{table}[tp]
\centering \small
\begin{tabular}{@{$\,$}l@{$\;\;$}c@{$\;$}c@{$\;$}c@{$\;$}|@{$\;$}c@{$\;$}c@{$\;$}c@{$\;$}|@{$\;$}c@{$\;$}c@{$\;$}c@{$\,$}}
& \multicolumn{3}{c}{\scriptsize\textsc{Trade-Off}} & \multicolumn{3}{c}{\scriptsize\textsc{Not-a-Trade-Off}} & \multicolumn{3}{c}{\scriptsize\textsc{Arg-Modifier}}        \\
                & P     & R     & \multicolumn{1}{c}{F1}    & P     & R     & \multicolumn{1}{c}{F1}    &P      & R     & F1 \\ \hline
\textit{dev}    & 76.92 & 74.07 & 75.47 & 75.45 & 81.01 & 78.13 & 53.60 & 47.18 & 50.19 \\
\textit{test}   & 89.71 & 84.72 & 87.14 & 72.08 & 72.08 & 72.08 & 52.05 & 57.58 & 54.68 
\end{tabular}
\caption{Overview of \textsc{SciIE} results for each of the relations in \textsc{FOBIE}. Note that the amount of gold relations for the development (dev) and test sets are not equal.}
\label{tab:rel_results}
\end{table}

A random sample of 250 sentences (16.1\%), which were annotated by using the RBS, was re-annotated by a second domain-expert. The inter-annotator agreement Cohen \textit{k} for both relations and span boundaries was found to be 92.93.

\subsection{Dataset description}
Table~\ref{tab:dataset_description} provides an overview of the statistics on \textsc{FOBIE}. The percentage of singleton keyphrases in \textsc{FOBIE} is only 20.4\%, a considerable difference with regard to \textsc{ScienceIE} (31\%) and \textsc{ACL RD-TEC 2.0} (83\%) \cite{Augenstein2017a}. The number of single token entities in \textsc{SciERC} is (31\%), with an average token length of 2.36 in comparison to 3.46 in \textsc{FOBIE}. The reason is that arguments of trade-offs are often phrases or Multi-Word Expressions (MWE), such as `\textit{immune response}'.

In \textsc{FOBIE} only 17.4\% of the sentences is shorter than 25 tokens, while in \textsc{SciERC} the average sentence length is 24.31 tokens. The much longer sentences in \textsc{FOBIE} is influenced by the presence of citations, but in general the sentences in full-text documents are longer. 
The final dataset comprises 1548 single sentences taken from 1292 unique documents and is randomly split into 1248 training, 150 development and 150 test instances. There is no source document overlap between the training, development and test set.

\section{Baseline Evaluation}
We train a state-of-the-art scientific RE system on \textsc{FOBIE} called \textsc{SciIE} \cite{Luan2018a}. The \textsc{SciIE} model is developed for the \textsc{SciERC} dataset to jointly extract the entities, relations and coreference annotations \cite{Luan2018a}. Furthermore, the model achieves state-of-the-art performance on the \textsc{ScienceIE} dataset, which includes the tasks of span identification and keyphrase extraction. For \textsc{FOBIE} we only train \textsc{SciIE} on entities and relations, where the entities are our annotated keyphrases; trigger spans and argument spans. 

\paragraph{\textsc{SciIE}} is a span-based IE model that takes as input an unlabeled sentence, e.g., no POS-tags or dependency labels have to be provided. All sequences of consecutive words in the sentence -- up to a given length -- are considered as a candidate span, while only a few represent the gold annotated keyphrases. Each token is represented as the concatenation of its word, character and ELMo embedding \cite{Peters2018}. Each candidate span is represented as a concatenation of the bi-LSTM outputs for the first token in the span and the last token in the span, as well as an attention mechanism over the span that is thought to represent the syntactic head of span \cite{Lee2017}. Span classification determines whether a span is actually a span, where a dummy non-span class is assigned to incorrect candidate spans. Pairs of candidate spans are fed to the relation classifier, where a dummy non-relation class is introduced. This multi-task setup enables weight sharing that improves performance on both tasks. See \cite{Luan2018a} for more details on \textsc{SciIE}.

We compare \textsc{RBS} and \textsc{SciIE} on detecting the correct boundaries of trigger words and argument phrases, see Table~\ref{tab:results}. As expected the neural approach outperforms the simple \textsc{RBS} by a large margin. We also evaluate how well \textsc{SciIE} performs on FOBIE with regards to extracting relations. The Relation Extraction (RE) setting is evaluated as a joint task -- the presence of spans has to be predicted correctly, as well as the relation types that may hold between them. We do not change the default hyper parameters, apart from setting the maximum span length to 14 tokens. This is required to deal with the longer keyphrases in FOBIE. 

Table~\ref{tab:rel_results} provides RE results per relation type. The model performs better on the test set with regards to extracting \textsc{Trade-Off} relations. Due to the relatively large amount of \textsc{Not-A-Trade-Off} (54.05\%) and \textsc{Argument-Modifier} (29.73\%) relations this difference becomes negligible in the overall RE setting, also see Table~\ref{tab:dataset_description}. The model performs notably worse on \textsc{Argument-Modifier} relations. This is likely the result of the large syntactic variety in argument modification.

\section{Conclusions}
We presented \textsc{FOBIE}, the first scientific IE dataset that focuses on the domain of biomimetics. \textsc{FOBIE} supports the extraction of \textsc{Trade-Offs} and syntactically similar relations from scientific biological texts. These relations enable cross-domain discovery of relevant scientific literature during biomimetics. However, the manners in which trade-off are expressed, and their arguments modified, varies a lot. This large syntactic variation and the long sentences in scientific text decrease the accuracy of extraction rules. Therefore, Rule-Based Systems (RBS) may not be able to deal with the long tail of patterns. The alternative machine learning approach requires time-consuming manual annotation of datasets. Combining the two approaches, a simple RBS can speed up the annotation of a small dataset by a domain expert considerably. The use of pretrained embeddings provides reliable flexibility with regards to syntactic variance.  

The manual annotation of \textsc{FOBIE} instances enables the training of neural Relation Extraction systems. The size of \textsc{FOBIE} is comparable to existing dataset for scientific Information Extraction. Unlike existing datasets the keyphrases in \textsc{FOBIE} are not classified into entity types.
We make FOBIE and the annotation guidelines publicly available at \url{https://github.com/rubenkruiper/FOBIE}.



\section*{Acknowledgments}
The authors would like to thank the reviewers for their useful comments, and gratefully acknowledge the financial support of the Engineering and Physical Sciences Research Council (EPSRC) Centre for Doctoral Training in Embedded Intelligence under grant reference EP/L014998/1 and the EPSRC Innovation Placement fund.

\section{Bibliographical References}
\label{main:ref}

\bibliographystyle{lrec}
\bibliography{Mendeley}


\end{document}